\documentclass[letterpaper, 10 pt, conference]{ieeeconf}

\IEEEoverridecommandlockouts
\usepackage{amsmath, amssymb}
\usepackage{booktabs}
\usepackage{multirow}
\usepackage{graphicx}
\usepackage{xcolor}
\usepackage{algorithm}
\usepackage{algorithmic}
\usepackage{listings}
\usepackage{newfloat}
\usepackage{microtype}
\usepackage[hyphens]{url}
\usepackage{cite}
\usepackage[hidelinks]{hyperref}

\floatstyle{ruled}
\newfloat{listing}{tb}{lst}
\floatname{listing}{Listing}

\title{\LARGE \bf
T2M Mamba: Motion Periodicity-Saliency Coupling Approach \\
for Stable Text-Driven Motion Generation
}

\author{Xingzu Zhan, Chen Xie, Honghang Chen, Yixun Lin, and Xiaochun Mai
\thanks{Xingzu Zhan, Chen Xie, Honghang Chen are with Shenzhen University, Shenzhen, China. Xiaochun Mai is with the College of Electronics and Information Engineering, Shenzhen University, Shenzhen, China, with the Shenzhen Key Laboratory of Digital Creative Technology, with the Guangdong Province Engineering Laboratory for Digital Creative Technology, and also with the Guangdong-Hong Kong Joint Laboratory for Big Data Imaging and Communication, Shenzhen, China.}%
\thanks{Yixun Lin is with Jinan University, Guangzhou, China.}%
\thanks{\emph{Corresponding author: Xiaochun Mai (e-mail: { maixc@szu.edu.cn}).}}%
}

\begin{document}

\maketitle
\thispagestyle{empty}
\pagestyle{empty}

%%%%%%%%%%%%%%%%%%%%%%%%%%%%%%%%%%%%%%%%%%%%%%%%%%%%%%%%%%%%%%%%%%%%%%%%%%%%%%%%
\begin{abstract}

Text-to-motion generation, which converts motion language descriptions into coherent 3D human motion sequences, has attracted increasing attention in fields, such as avatar animation and humanoid robotic interaction. Though existing models have achieved significant fidelity, they still suffer from two core limitations: \textbf{(i)} They treat motion periodicity and keyframe saliency as independent factors, overlooking their coupling and causing generation drift in long sequences. \textbf{(ii)} They are fragile to semantically equivalent paraphrases, where minor synonym substitutions distort textual embeddings, propagating through the decoder and producing unstable or erroneous motions. In this work, we propose T2M Mamba to address these limitations by \textbf{(i)} proposing Periodicity-Saliency Aware Mamba, which utilizes novel algorithms for keyframe weight estimation via enhanced Density Peaks Clustering and motion periodicity estimation via FFT-accelerated autocorrelation to capture coupled dynamics with minimal computational overhead, and \textbf{(ii)} constructing a Periodic Differential Cross-modal Alignment Module (PDCAM) to enhance robust alignment of textual and motion embeddings. Extensive experiments on HumanML3D and KIT-ML datasets have been conducted,  confirming the effectiveness of our approach, achieving an FID of 0.068 and consistent gains on all other metrics.

\end{abstract}

%%%%%%%%%%%%%%%%%%%%%%%%%%%%%%%%%%%%%%%%%%%%%%%%%%%%%%%%%%%%%%%%%%%%%%%%%%%%%%%%
\section{Introduction}
Text-to-motion generation is a fundamental task in various fields including avatar animation, humanoid robots, and gaming, which transforms a natural-language description about motion into a semantically coherent and physically plausible high-dimensional temporal sequence of human poses.

Text‑to‑motion generation has evolved from early latent‑regression networks, through high‑fidelity yet costly diffusion transformers, to linear‑time Mamba state‑space models and finally tokenised motion–language frameworks that exploit large‑scale language modelling for long horizon synthesis~\cite{zhan2025multi,guo2022generating,zhang2024motiondiffuse,tevet2022human,barquero2024seamless,zhang2024motion,zhang2023t2mgptgeneratinghumanmotion}.
Building on these technological advances, researchers have begun to explore ways of enhancing the quality of generated motion. MMDM~\cite{chen2024text} and MoMask~\cite{guo2024momask} boost quality by reconstructing randomly masked frames, whereas KMM~\cite{zhang2024kmm} and KeyMotion~\cite{geng2024textguided3dhumanmotion} mask only algorithmically detected keyframes to focus the model on motion pivots. Researchers have also begun to exploit motion periodicity. For example, DiffusionPhase~\cite{wan2023diffusionphasemotiondiffusionfrequency} and FlowMDM~\cite{barquero2024seamless} introduce frequency or phase domain representations to help motion generation. We can see that existing work often overlooks the coupling of keyframes and motion periodicity. However, in fact that keyframe learning and cycle control are not isolated. A single motion sequence often contains multiple motion patterns, which may exhibit periodicity, e.g., running or walking, with keyframes potentially serving as boundaries between these patterns. 
Besides, recent work SOTA~\cite{chen_2024} indicates that current models are highly sensitive to semantically equivalent paraphrases, even minor synonym substitutions can trigger large drifts in textual embeddings. Subsequently, the drifts will be progressively amplified during decoding, resulting in divergent or erroneous motion sequences. The main reason for the low stability of textual and motion mapping is that the cross-attention suffers from the time scale mismatch, that is, one word corresponds to multiple motion frames, or multiple words describe an instantaneous action. Hence, motivated by the issues analyzed above, this paper aims to i) \textit{explore the coupling relationship between keyframes and motion periodicity}, then exploit the coupling to \textit{eliminate the mamba's history forgetting} in the long sequence motion generation, and ii) \textit{tackle the time scale mismatch issue} for enhancing the consistency and stability of the text-to-motion mapping.

To this end, we introduce \emph{T2M Mamba}, a novel methodology for human motion generation that comprises two principal modules: \emph{Periodicity-Saliency Aware Mamba} and the \emph{Periodic Differential Cross-modal Alignment Module (PDCAM)}. To address the first issue, T2M\,Mamba detects physically meaningful turning frames, measures their saliency to obtain keyframe weights. Then T2M\,Mamba applies an FFT-accelerated autocorrelation function between successive keyframes to estimate the motion period; Periodicity-Saliency Aware\,Mamba subsequently modifies Mamba’s input projection matrix with these keyframe weights, alleviates Mamba’s historical forgetting, and injects the estimated period to capture the latent motion rhythm. To confront the second issue, we propose the Periodic Differential Cross-modal Alignment Module (PDCAM). PDCAM integrates keyframes to pinpoint pivotal motion segments. It leverages motion periodicity to exploit repetitive patterns, enabling consistent alignment. Furthermore, it incorporates differential essence to emphasize semantic differences while mitigating noise. As a result, the module reliably extracts key information and semantics. This yields robust cross-modal alignment and enhanced stability across varying time scales.
Overall, our contributions can be summarised as follows.
\begin{itemize}
\item We propose novel algorithms for keyframe weight estimation and motion periodicity estimation, utilizing enhanced density peaks clustering (DPC) for adaptive keyframe detection within motion segments and FFT-accelerated autocorrelation for identifying dominant periods in motion signals, all while incurring almost no additional computational overhead.

\item We introduce Periodicity-Saliency Aware Mamba, which combines keyframe weighting and motion periodicity modeling to alleviate historical forgetting in long sequences, and learn coordinated motion patterns. Furthermore, we propose the Periodic Differential Cross-modal Alignment Module (PDCAM) to bolster robust cross-modal mapping against textual perturbations by integrating keyframe importance and motion phase encoding.

\item We conduct extensive experiments on the HumanML3D and KIT-ML datasets to validate the effectiveness of T2M Mamba. We achieve substantial improvements across all metrics, including a notable FID score of 0.068, demonstrating the effectiveness and efficiency of our proposed approach.
\end{itemize}

\section{Related Work}
\subsection{Text-to-Motion Generation}
Text-to-motion generation has evolved through VAE-based, diffusion-based, and hybrid methods. VAE-based approaches, such as Action2Motion~\cite{Guo_2020} and TM2T~\cite{guo2022tm2tstochastictokenizedmodeling}, model motion via latent variables to enable probabilistic sampling, promoting diversity and interpretability in continuous or discrete spaces. However, they often produce unrealistic results and suffer from mode collapse due to poor generalization. Diffusion-based methods, like MotionDiffuse~\cite{zhang2024motiondiffuse} and MDM~\cite{tevet2022human}, iteratively denoise noisy motion sequences conditioned on text, excelling as motion learners for multi-modal conditioning and high-fidelity outputs, but they are limited in long sequences due to generation drift and high computational demands. Hybrid models, such as MLD~\cite{chen2023executing} and AttT2M~\cite{zhong2023attt2m}, combine VAEs for latent compression with diffusion for generation, struggling with fine-grained details and high computational complexity. Based on diffusion framework,
our T2M Mamba uniquely couples motion periodicity and keyframe saliency, addressing generation drift in long sequences. Furthermore, it introduces the Periodic Differential Cross-modal Alignment Module (PDCAM) to robustly align textual and motion embeddings, mitigating instability from synonym substitutions.

\subsection{State Space Models}
State Space Models (SSMs) describe dynamic systems through intermediate state variables. S4~\cite{gu2022efficientlymodelinglongsequences} introduce structured low-rank corrections for stable diagonalization, enabling efficient long-sequence modeling with linear complexity. Based on S4, Mamba~\cite{gu2023mamba} incorporates a selective mechanism and hardware-aware algorithms, parameterizing SSMs based on input sequences to achieve content-aware processing while maintaining efficiency. 
% In the visual domain,Vim~\cite{zhu2024visionmambaefficientvisual} uses bidirectional SSMs with positional embeddings for vision tasks, while VMamba~\cite{liu2024vmambavisualstatespace} introduces cross-scan modules to handle non-causal 2D data, outperforming CNNs~\cite{kim2014convolutional} and Transformers~\cite{vaswani2017attention} on benchmarks like ImageNet. 
% Other extensions, such as EfficientVMamba~\cite{pei2025efficientvmamba} and LocalMamba~\cite{huang2024localmamba}, optimize scanning strategies for low-level vision, achieving state-of-the-art results in denoising and super-resolution with reduced parameters.
In motion generation, Motion Mamba~\cite{zhang2024motion},  TM-Mamba~\cite{wang2024text} and Infinimotion~\cite{zhang2024infinimotion} offer linear-time complexity for long-sequence modeling via selective state updates, but they suffer from historical forgetting in extended sequences, where early inputs lose influence, and overlook coupled dynamics like periodicity and saliency in motion data.
By comparison, our Periodicity-Saliency Aware Mamba explicitly considers keyframe importance and injects estimated periodicity into Mamba's sequences, alleviating historical forgetting more effectively than standard SSMs. By integrating these coupled factors with minimal overhead, it captures latent motion rhythms and coordinated patterns absent in prior SSMs.

\begin{figure*}[t] % t = 页面顶部
    \centering
    % 先在图形软件中裁剪，避免在 includegraphics 中使用 trim/clip
    \includegraphics[width=0.93\textwidth]{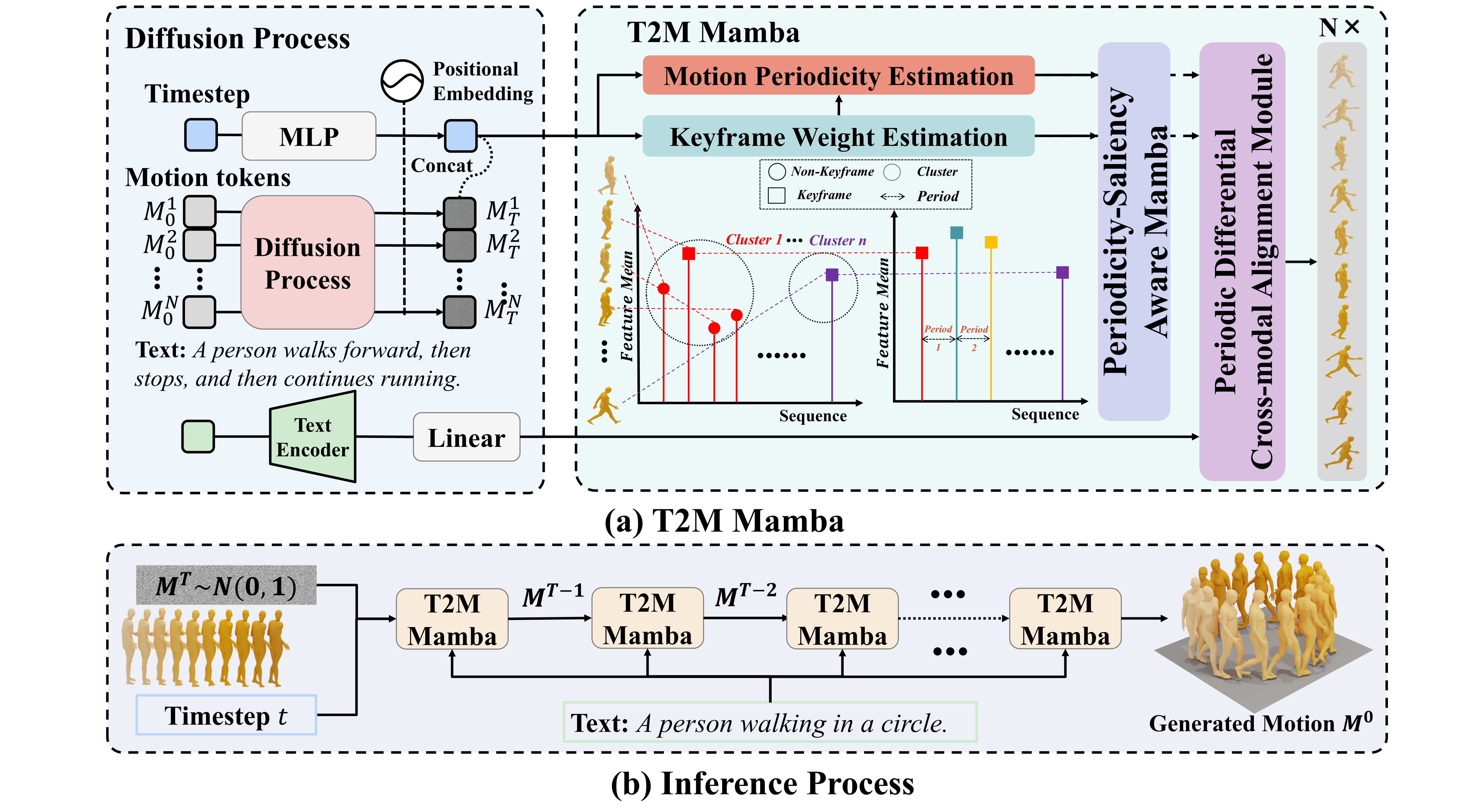}
    \vspace{-1.5em}
    \caption{\textbf{The overview of the proposed T2M Mamba.} \textbf{(a) T2M Mamba.} Our T2M Mamba consisting of N basic blocks aims to predict clean motion sequence \textbf{(b) Inference Process.} Starting from Gaussian noise, the model iteratively denoises to generate a clean motion sequence $ M^0$ semantically aligned with the input text prompt.} 
    \vspace{-4mm}
    \label{T2MMamba}
\end{figure*}
\section{Methodology}
\subsection{Overview of T2M Mamba}
To address the challenges of overlooked coupling between keyframe saliency and motion periodicity, as well as instability in text-to-motion mapping under semantic paraphrases, we propose T2M Mamba, as shown in Fig~\ref{T2MMamba}, a framework that integrates Periodicity-Saliency Aware Mamba and Periodic Differential Cross-modal Alignment Module (PDCAM). Given a noisy motion sequence after $T$ diffusion steps,
$\{M_T^{1}, M_T^{2}, M_T^{3}, \dots, M_T^{N}\}$,
T2M Mamba conditions on the CLIP-encoded text prompt and the current diffusion
timestep~$T$ to predict the corresponding clean motion sequence
$\{M_0^{1}, M_0^{2}, M_0^{3}, \dots, M_0^{N}\}$.

\subsection{Motion Periodicity-Saliency Coupling}
In text-to-motion generation, capturing keyframe saliency and inherent periodicity is crucial for maintaining physical plausibility and rhythmic coherence. We first introduce methods for keyframe weight estimation and motion periodicity estimation. These estimations serve as foundational cues for the Periodicity-Saliency Aware Mamba and Periodic Differential Cross-modal Alignment Module, as detailed in the following subsections.

\paragraph{keyframe weight estimation} To estimate keyframe weights, we partition the motion stream into \(N\) equal-length segments and apply an enhanced Density Peaks Clustering (DPC) within each segment to automatically detect keyframes based on local density and separation distance, followed by assigning normalized weights to emphasize their importance. Specifically, we first partition the motion stream into $N$ equal-length temporal segments $\{X_1,\dots ,X_N\}$. Within each segment, an improved \emph{Density Peaks Clustering} (DPC)~\cite{tang2022deepunsupervisedkeyframe} is performed.  
Given a segment $\{\mathbf x_i\}_{i=1}^{M}$, pairwise similarity is measured by the Euclidean distance $d_{ij}= \lVert \mathbf x_i - \mathbf x_j \rVert$, the subscript $j$ denotes the index of a sample distinct from $i$ within the current segment, paired with $i$ to compute the Euclidean distance $d_{ij}$.
And the cut-off distance $d_c$ is selected adaptively so that approximately $1$–$2\%$ of the points lie within this neighbourhood.  
The local density is defined by

\begin{equation}
\rho_i = \sum_{j\neq i} e^{-\left(\frac{d_{ij}}{d_c}\right)^{\!2}}.
\end{equation}

After computing the local density $\rho_i$ for each sample, relying solely on density magnitude is insufficient, as a high-density region may contain multiple peaks that could be misidentified as centers without distinction. To address this, Density Peaks Clustering introduces the minimum separation distance $\delta_i$, defined as the smallest Euclidean distance to any point with higher density (or the maximum distance if none exists), i.e.,
\vspace{-0.2em}
\begin{equation}
\delta_i = 
\begin{cases}
\displaystyle
\min_{j:\,\rho_j>\rho_i} d_{ij}, & \text{if } \exists\,j:\rho_j>\rho_i,\\[6pt]
\displaystyle
\max_{j} d_{ij}, & \text{otherwise}.
\end{cases}
\end{equation}
\vspace{-0.2em}
The composite peak score is $\gamma_i = \rho_i \delta_i$, ensuring candidate centers are dense (high $\rho_i$) and well-separated (large $\delta_i$). After computing all $\gamma_i$, we sort them descendingly and use the elbow point in the $\gamma$ curve to infer the number of keyframes automatically.
We then construct the keyframe weight matrix $\mathbf{M} \in \mathbb{R}^{L \times 1}$ by concatenating results from the $N$ segments in temporal order, where for each frame $i$,
\begin{equation}
\mathbf{M}_i = 
\begin{cases}
\bar{\gamma}_i, & i \in \mathcal{K} \\
1, & \text{otherwise},
\end{cases}
\quad \bar{\gamma}_i = \frac{\gamma_i - \gamma_{\min}}{\gamma_{\max} - \gamma_{\min}},
\label{key}
\end{equation}
with $\mathcal{K}$ denoting the set of detected keyframes, and $\gamma_{\min}$, $\gamma_{\max}$ the minimum and maximum $\gamma$ values among detected keyframes in the sequence.

\paragraph{Motion Periodicity Estimation}
To estimate motion periodicity, we segment the motion sequence into intervals between consecutive keyframes and independently analyze each segment's one-dimensional motion signal using FFT-accelerated autocorrelation to detect dominant periods, determining periodicity based on peak, prominence, and spectral entropy criteria. For each segment $s$ with length $L_s$, we extract the signal $\mathbf{x}_s \in \mathbb{R}^{L_s}$ by averaging across feature dimensions and compute its normalized autocorrelation function (ACF) via the Wiener-Khinchin theorem rather than directly evaluating the ACF for efficiency.
\begin{equation}
\mathbf{x}_s = \frac{1}{D}\sum_{d=1}^{D} \mathbf{F}_{d}, \quad
R(\tau) = \frac{1}{R(0)} \cdot \mathcal{F}^{-1}\{|\mathcal{F}\{\mathbf{x}_s\}|^2\},
\end{equation}
where $\mathbf{F}_d$ is the $d$-th feature dimension, $\mathcal{F}$ and $\mathcal{F}^{-1}$ denote the Fast Fourier Transform and its inverse, $\tau$ is the time lag measuring signal shift, and $R(0)$ (the signal's variance) normalizes $R(\tau)$ to $[-1, 1]$ for similarity assessment at various lags.

We identify the dominant period by locating the first significant peak in the autocorrelation function $R(\tau)$ for time lag $\tau > 0$. We employ the peak ratio criterion~\cite{wen2023robustdominantperiodicitydetection}, prominence criterion~\cite{Ravbar2021}, and spectral entropy criterion~\cite{Huang2022} to assess periodicity in motion segments. Stringent thresholds across these criteria effectively filter out spurious patterns, preventing erroneous assignment of periodicity to inherently non-periodic motions. A segment is classified as periodic if it satisfies three criteria,
\begin{equation}
\begin{cases}
R(\tau_{\max}) > \theta_{\text{peak}} & \text{(peak ratio)} \\
R(\tau_{\max}) - \bar{R} > \theta_{\text{prom}} & \text{(prominence)} \\
H_{\text{spec}} < \theta_{\text{ent}} & \text{(spectral entropy),}
\end{cases}
\end{equation}
where $\tau_{\max} = \arg\max_{\tau>0} R(\tau)$ denotes the lag position of the maximum peak, $R(\tau_{\max})$ is the autocorrelation value at the peak position, $\bar{R}$ represents the mean value of the autocorrelation function,$H_{\text{spec}}$ is the spectral entropy which quantifies the regularity/randomness of the power spectrum. $\theta_{\text{peak}}$, $\theta_{\text{prom}}$, and $\theta_{\text{ent}}$ are thresholds for peak ratio, prominence, and spectral entropy, respectively. The spectral entropy is calculated as
\vspace{-2mm}
\begin{equation}
H_{\text{spec}} = -\frac{1}{\log N} \sum_{k=0}^{N-1} p_k \log p_k, \quad p_k = \frac{|X_k|^2}{\sum_{j=0}^{N-1} |X_j|^2},
\end{equation}
where $X_k$ denotes the $k$-th frequency component of the Fourier transform, $p_k$ is the normalized power spectrum representing the relative power at frequency $k$, and $N$ is the FFT size. The normalization factor $1/\log N$ ensures $H_{\text{spec}} \in [0, 1]$.

For periodic segments, the detected period is $T = \tau_{\max}$. Otherwise, the period T is set as the segment length, $T = L_s$. Finally, we compute the instantaneous phase for each frame, i.e.,
\begin{equation}
\label{phase}
\phi(t) = \frac{2\pi t}{T}, \quad \mathbf{\Phi} = [\sin\phi(t), \cos\phi(t)].
\end{equation}
By concatenating the obtained $\phi(t)$ and $\mathbf{\Phi}$ along the time order, we form the radian vector $\phi \in \mathbb{R}^{L \times 1}$, utilized in the Periodic Differential Cross‑modal Alignment Module (subsection~\ref{PDCAM}), and the phase encoding matrix $\mathbf{\Phi} \in \mathbb{R}^{L \times 2}$, employed in the Periodicity-Saliency Aware Mamba module (subsection~\ref{Periodicity-Saliency Aware Mamba}).

\begin{figure}[tb]  % t=top, b=bottom；不建议用 h(ere) 以免破版
  \centering
  \includegraphics[width=0.90\columnwidth]{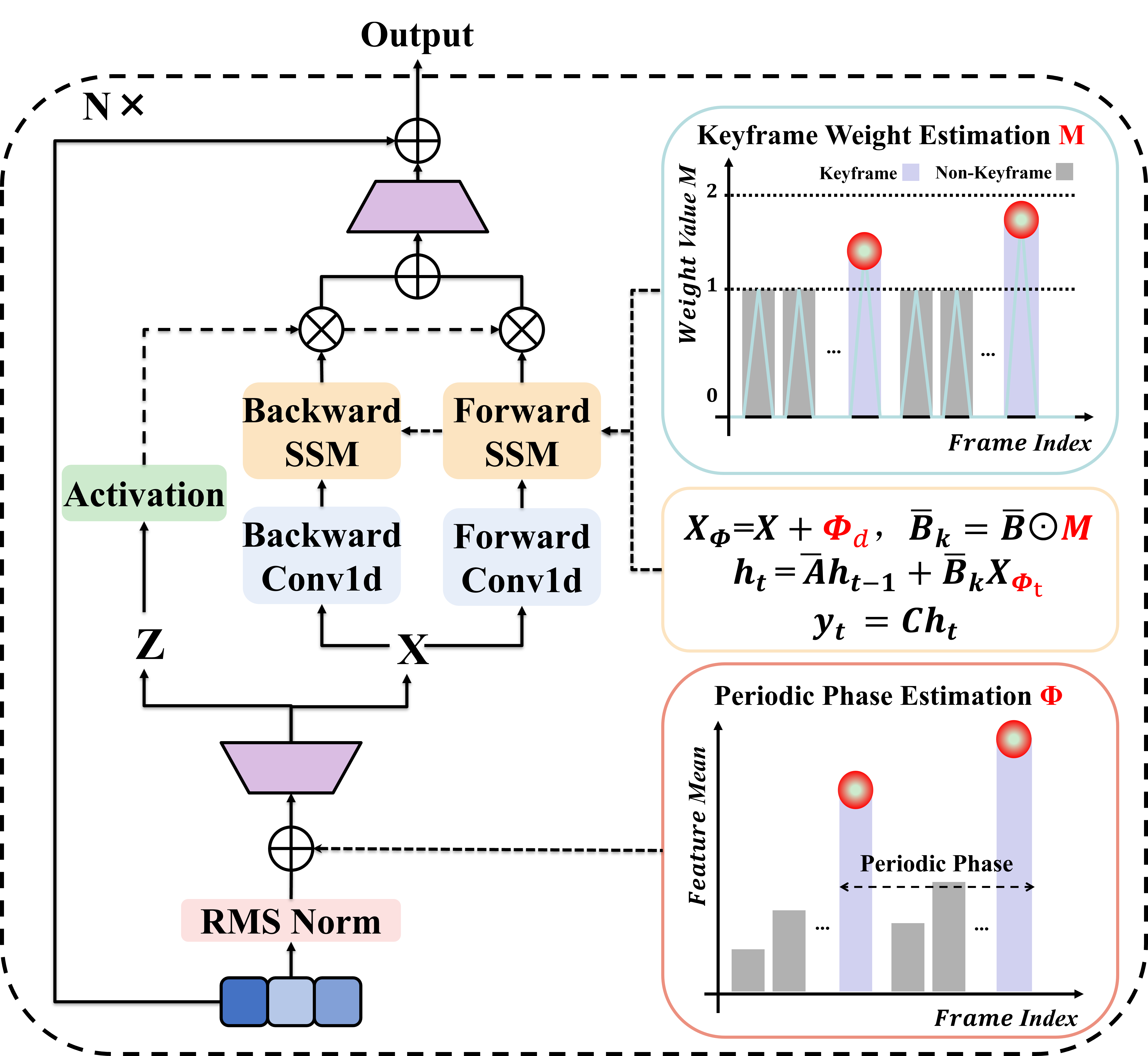}  % 单栏宽度
  \vspace{-0.5em}
  \caption{Illustration of our Periodicity-Saliency Aware Mamba. $\odot$ denotes dot product.
  }
  \label{fig.attention}
  \vspace{-1em}  % （可选）紧凑排版，按需要微调
  
  \label{mamba}
\end{figure}

\subsection{Periodicity-Saliency Aware Mamba}
\label{Periodicity-Saliency Aware Mamba}

Due to the state-space recurrence in Mamba~\cite{gu2023mamba}, which continuously compresses historical information into the hidden state $h(t)$, physically critical frames become severely attenuated, and the model fails to capture latent periodicity in cyclic motions such as walking, running, and dancing, resulting in beat drift and pose jitter. To mitigate these limitations, we propose Periodicity-Saliency Aware Mamba, as shown in Fig~\ref{mamba} which leverages the detected keyframe weights and periodicity estimates from the previous section. Specifically, it constructs a saliency matrix from the keyframe weights to replace the input projection $\bar{B}$ of Mamba, explicitly amplifying the influence of important frames; meanwhile, it injects sine-cosine phase encodings derived from the estimated periods at every timestep. This dual strategy simultaneously reinforces keyframe supervision and enforces global rhythmic coherence, thereby addressing deficiencies caused by information compression. We detail the Periodicity-Saliency Aware Mamba formulation below.

Given a motion sequence ${X} \in \mathbb{R}^{L \times D}$, a keyframe weight matrix ${M} \in \mathbb{R}^{L \times 1}$, and a phase encoding matrix ${\Phi} \in \mathbb{R}^{L \times 2}$, we first linearly project the phase encoding matrix to ${\Phi}_d \in \mathbb{R}^{L \times D}$ via $\Phi_d = \Phi W_\phi$, where $W_\phi \in \mathbb{R}^{2 \times D}$ is a learnable projection matrix. The input sequence \({X}\) is then added to \({\Phi}_d\) to obtain an explicitly rhythm-enhanced input \({X}_\Phi\). To enable Mamba to \emph{explicitly increase attention on keyframes}, we multiply the obtained keyframe weight matrix \({F}\) with the input projection matrix \(\bar{B} \in \mathbb{R}^{N \times 1}\) in Mamba's state space equation, yielding \(\bar{{B}_k}\). This allows more keyframe information to be stored in the historical state \({h}_t\). The overall formulation of Periodicity-Saliency Aware Mamba is thus given by
\vspace{-0.2em}
\begin{equation}
\begin{cases}
{X}_\Phi = {X} + {\Phi}_d, \;\;
\bar{{B}_k} = \bar{{B}} \odot {M},\\
h_t = \bar{A} h_{t-1} + \bar{{B}_k} {{X}_\Phi}_t, \\
y_t = C h_t.
\end{cases}
\end{equation}
Where \(\odot\) denotes element-wise multiplication. Note that the keyframe detection and periodicity detection modules are computationally efficient in the training, our Periodicity-Saliency Aware Mamba achieves explicit attention to keyframes and learning of periodic rhythms with almost no additional computational time overhead.

Periodicity-Saliency Aware Mamba maintains computational complexity comparable to standard Mamba and preserves its linear time complexity while addressing limitations in long-sequence motion generation. It integrates explicit keyframe weighting and motion phase encoding to mitigate historical forgetting from information compression and to capture rhythmic patterns in periodic actions.

\subsection{Periodic Differential Cross‑modal Alignment Module}

\label{PDCAM}

\begin{figure}[tb]  % t=top, b=bottom；不建议用 h(ere) 以免破版
  \centering
  \includegraphics[width=0.80\columnwidth]{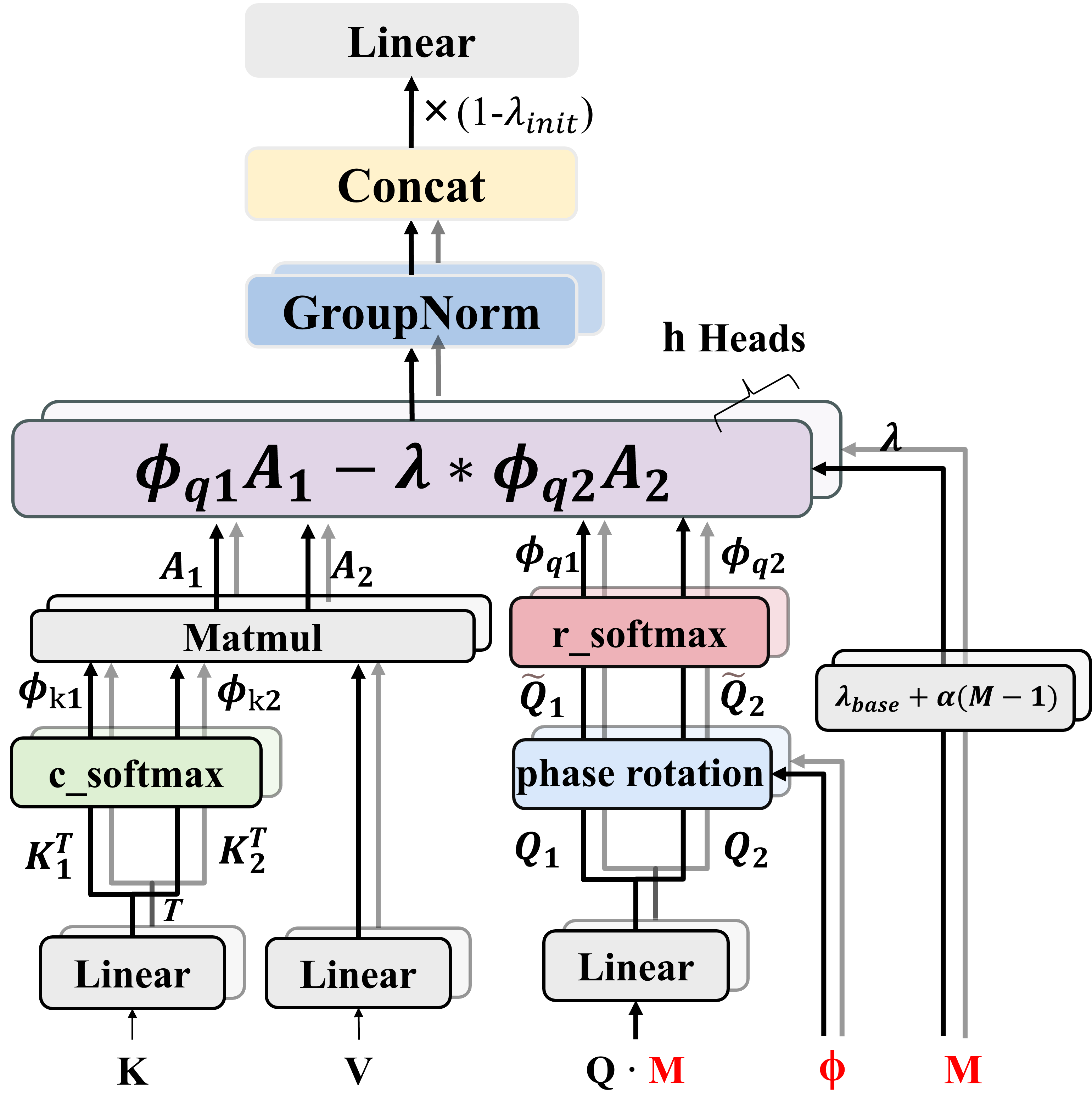}  % 单栏宽度
    \vspace{-0.5em}  % （可选）紧凑排版，按需要微调
  \caption{Illustration of the proposed PDCAM pipeline.%
  % 若需要，可在句末加引号或引用：\cite{yourBibKey}.
  }

  \label{fig:attention}
\end{figure}

% Defining the document class and necessary packages
In text-to-motion generation, cross-attention often suffers from time-scale mismatch, where one word maps to multiple motion frames or multiple words describe an instantaneous action, leading to unstable cross-modal alignment. To address this issue, we improve the differential attention from the Differential Transformer~\cite{ye2025differentialtransformer}, by using keyframes, motion periodicity, and differential essence to extract key information and semantics, thereby achieving robust alignment and enhancing cross-modal stability.

Unlike the Differential Transformer, which builds two softmax attention maps on the same query–key pair, PDCAM constructs two linear‑attention maps from phase‑rotated query halves and value, then subtracts them. PDCAM yielding a lightweight differential operator that accentuates discriminative cross‑modal cues, acting as a differential amplifier. Additionally, PDCAM uses keyframe weight matrix $M$ from Eq.~\ref{key} and radian matrix $\phi$ from Eq.~\ref{phase} to enhance the generating effects of motion sequences. Specifically, given a motion sequence $X \in \mathbb{R}^{L_\mathrm{m} \times D}$ and a textual sequence $T \in \mathbb{R}^{L_\mathrm{t} \times D}$, we first compute the projections,
\begin{align}
[Q_1; Q_2] = (X W^{Q}) \cdot M,\; \;
[K_1;K_2] = T W^{K},\;
V = T W^{V},
\end{align}
where $W^{Q}, W^{K} $, $W^{V} \in \mathbb{R}^{D \times 2D}$ are learnable projection matrices, and keyframe weight matrix $M$ is applied element-wise along the sequence dimension

Additionally, to enhance Coordination of motion sequences, PDCAM utilize a radian matrix $\phi \in \mathbb{R}^{L}$ from Eq.~\ref{phase}, where $\phi_i \in [0, 2\pi)$ represents the phase of the $i$-th frame. A learnable phase temperature $\beta$ modulates the phase rotation applied to the query slices, i.e.,
\begin{alignat}{2}
\tilde{{Q}}_1 = Q_1 \cos(\beta \phi) - Q_2 \sin(\beta \phi), \notag \\
\quad \tilde{{Q}}_2 = Q_1 \sin(\beta \phi) + Q_2 \cos(\beta \phi),
\end{alignat}
where $\cos(\beta \phi)$ and $\sin(\beta \phi)$ are applied element-wise, and $\tilde{{Q}_1} , \tilde{{Q}}_2  \in \mathbb{R}^{L_\mathrm{m} \times D}$ are the rotated query slices. This rotation injects a periodic inductive bias into the query representations, enabling the model to better capture relative positional dependencies and rhythmic patterns in motion sequences. The key slices $K_1, K_2$ remain unrotated to preserve textual context integrity.

The attention scores are computed with column-wise softmax for queries and row-wise softmax for keys,
\begin{alignat}{2}
\Phi_{q1} = \mathrm{softmax}_{\text{col}}(\tilde{{Q}}_1), \quad \Phi_{q2} = \mathrm{softmax}_{\text{col}}(\tilde{{Q}}_2), \notag \\
\Phi_{k1} = \mathrm{softmax}_{\text{row}}(K_1), \quad \Phi_{k2} = \mathrm{softmax}_{\text{row}}(K_2).
\end{alignat}
The differential attention operator is defined as
\begin{equation}
\begin{cases}
A_1 = \Phi_{k1}^{\top} V, \quad A_2 = \Phi_{k2}^{\top} V, \\
\lambda_{\text{base}} = \exp\left(\lambda_{q1} \lambda_{k1}\right) - \exp\left(\lambda_{q2} \lambda_{k2}\right) + \lambda_{\mathrm{init}}, \\
\lambda = \lambda_{\text{base}} \cdot \left(1 + \alpha_{\text{imp}} (M - 1)\right), \\
\operatorname{LinDiffCroAttn}(X, T) = \Phi_{q1} A_1 - \lambda \Phi_{q2} A_2,
\end{cases}
\end{equation}
where $\lambda_{q1}, \lambda_{q2}, \lambda_{k1}, \lambda_{k2} \in \mathbb{R}^{H \times D/H}$ are learnable parameters per head, $\lambda_{\mathrm{init}}$ is a scalar initialized to 0.8, and $\alpha_{\text{imp}}$ is a learnable parameter controlling the keyframe importance modulation. The token-wise $\lambda$ adapts the suppression strength dynamically based on keyframe importance.

To leverage multi-head attention~\cite{vaswani2017attention}, we assign each head $i$ (for $i = 1, \dots, H$) its own projection matrices $W_i^{Q}, W_i^{K}, W_i^{V}$, while sharing $\lambda_{\text{base}}$, $\alpha_{\text{imp}}$, and $\beta$ across heads. The $i$-th head and multi-head output are computed as
\begin{equation}
\begin{cases}
\text{Head}_i = \operatorname{LinDiffCroAttn}\left(X, T, W_i^{Q}, W_i^{K}, W_i^{V}, \lambda, M, \phi\right), \\
O = \operatorname{RMSNorm}\left(\operatorname{concat}(\text{Head}_1, \dots, \text{Head}H)\right) (1 - \lambda{\mathrm{init}}), \\
\operatorname{MultiHeadLinearDiffCroAttn}(X, T) = O W^{o},
\end{cases}
\end{equation}
where $W^{o} \in \mathbb{R}^{2D \times D}$ is a learnable output projection matrix. The keyframe importance weighting and phase-rotary encoding enhance temporal alignment, making PDCAM particularly effective for capturing dynamic motion patterns in text-to-motion generation.

\label{experiments}
\begin{table*}[ht]
    \centering
    \small
\resizebox{0.87\textwidth}{!}{%
    \begin{tabular}{l@{\hspace{0.3cm}}c@{\hspace{0.4cm}}c@{\hspace{0.4cm}}c@{\hspace{0.4cm}}c@{\hspace{0.4cm}}c@{\hspace{0.4cm}}c@{\hspace{0.4cm}}c}
        \toprule
        
        \multirow{2}{*}{\hspace{12mm} Methods} & \multicolumn{3}{c}{R Precision$\uparrow$} & \multirow{2}{*}{FID$\downarrow$} & \multirow{2}{*}{MM Dist$\downarrow$} & \multirow{2}{*}{Diversity$\rightarrow$} & \multirow{2}{*}{MModality$\uparrow$} \\
        \cmidrule(lr){2-4}
        & Top 1 & Top 2 & Top 3 & & & & \\
        \midrule
        \hspace{7mm}\textit{On the HumanML3D} & 
        \textit{0.511}$^{\pm .003}$ & 
        \textit{0.703}$^{\pm .003}$ & 
        \textit{0.797}$^{\pm .002}$ & 
        \textit{0.002}$^{\pm .000}$ & 
        \textit{2.974}$^{\pm .008}$ & 
        \textit{9.503}$^{\pm .065}$ & -- \\

        \midrule
        T2M~\cite{guo2022generating}  & 0.457$^{\pm .002}$ & 0.639$^{\pm .003}$ & 0.740$^{\pm .003}$ & 1.067$^{\pm .002}$ & 3.340$^{\pm .008}$ & 9.188$^{\pm .002}$ & 2.090$^{\pm .083}$ \\
        MDM~\cite{tevet2022human} & 0.320$^{\pm .005}$ & 0.498$^{\pm .004}$ & 0.611$^{\pm .007}$ & 0.544$^{\pm .044}$ & 5.566$^{\pm .027}$ & 9.559$^{\pm .086}$ & \textbf{2.799}$^{\pm .072}$ \\
        MotionDiffuse~\cite{zhang2024motiondiffuse} & 0.491$^{\pm .001}$ & 0.681$^{\pm .001}$ & 0.782$^{\pm .001}$ & 0.630$^{\pm .001}$ & 3.113$^{\pm .001}$ & 9.410$^{\pm .049}$ & 1.553$^{\pm .042}$ \\
        MLD~\cite{chen2023executing} & 0.481$^{\pm .003}$ & 0.673$^{\pm .003}$ & 0.772$^{\pm .002}$ & 0.473$^{\pm .013}$ & 3.196$^{\pm .010}$ & 9.724$^{\pm .082}$ & 2.413$^{\pm .079}$ \\
        Motion Mamba~\cite{zhang2024motion} & 0.502$^{\pm .003}$ & 0.693$^{\pm .002}$ & 0.792$^{\pm .002}$ & 0.281$^{\pm .009}$ & 3.060$^{\pm .058}$ & 9.871$^{\pm .084}$ & 2.294$^{\pm .058}$ \\
        T2M-GPT~\cite{zhang2023t2mgptgeneratinghumanmotion} & 0.492$^{\pm .003}$ & 0.679$^{\pm .002}$ & 0.775$^{\pm .002}$ & 0.141$^{\pm .005}$ & 3.121$^{\pm .009}$ & 9.722$^{\pm .082}$ & 1.831$^{\pm .048}$ \\            
        AttT2M~\cite{zhong2023attt2m} & 0.499$^{\pm .003}$ & 0.690$^{\pm .002}$ & 0.786$^{\pm .002}$ & 0.112$^{\pm .006}$ & 3.038$^{\pm .007}$ & 9.700$^{\pm .090}$ & 2.452$^{\pm .051}$ \\   
        MoMask~\cite{guo2024momask} & \textbf{0.521}$^{\pm .002}$ & \textbf{0.713}$^{\pm .002}$ & \textbf{0.807}$^{\pm .002}$ &  \textbf{0.045}$^{\pm .002}$ & \textbf{2.958}$^{\pm .008}$ & - & 1.241$^{\pm .040}$ \\  
        \midrule
        T2M Mamba (Ours) & \underline{0.506}$^{\pm .002}$ & \underline{0.696}$^{\pm .002}$ & \underline{0.793}$^{\pm .002}$ & \underline{0.068}$^{\pm .004}$ & \underline{3.034}$^{\pm .007}$ & \textbf{9.497}$^{\pm .010}$ & 2.310$^{\pm .065}$ 
        \\
        \midrule
        \hspace{9mm}\textit{On the KIT-ML} & 
        \textit{0.424}$^{\pm .005}$ & 
        \textit{0.649}$^{\pm .006}$ & 
        \textit{0.779}$^{\pm .006}$ & 
        \textit{0.031}$^{\pm .004}$ & 
        \textit{2.788}$^{\pm .012}$ & 
        \textit{11.08}$^{\pm .097}$ & -- \\
        \midrule
        T2M~\cite{guo2022generating} & 0.370$^{\pm .005}$ & 0.569$^{\pm .007}$ & 0.693$^{\pm .007}$ & 2.770$^{\pm .109}$ & 3.401$^{\pm .008}$ & 10.91$^{\pm .119}$ & 1.482$^{\pm .065}$ \\
        MDM ~\cite{tevet2022human} & 0.164$^{\pm .004}$ & 0.291$^{\pm .004}$ & 0.396$^{\pm .004}$ & 0.497$^{\pm .021}$ & 9.191$^{\pm .022}$ & 10.85$^{\pm .109}$ & 1.907$^{\pm .214}$ \\
        MotionDiffuse ~\cite{zhang2024motiondiffuse} & 0.417$^{\pm .004}$ & 0.621$^{\pm .004}$ & 0.739$^{\pm .004}$ & 1.954$^{\pm .062}$ & 2.958$^{\pm .005}$ & \textbf{11.10}$^{\pm .143}$ & 0.730$^{\pm .013}$ \\
        MLD ~\cite{chen2023executing} & 0.390$^{\pm .008}$ & 0.609$^{\pm .008}$ & 0.734$^{\pm .007}$ & 0.404$^{\pm .027}$ & 3.204$^{\pm .027}$ & 10.80$^{\pm .117}$ & \underline{2.192}$^{\pm .071}$ \\
        Motion Mamba~\cite{zhang2024motion} & 0.419$^{\pm .006}$ & 0.645$^{\pm .005}$ & 0.765$^{\pm .006}$ & 0.307$^{\pm .041}$ & 3.021$^{\pm .025}$ & \underline{11.02}$^{\pm .098}$ & 1.678$^{\pm .064}$ \\
        T2M-GPT~\cite{zhang2023t2mgptgeneratinghumanmotion} & 0.416$^{\pm .006}$ & 0.627$^{\pm .006}$ & 0.745$^{\pm .006}$ & 0.514$^{\pm .029}$ & 3.007$^{\pm .023}$ & 10.92$^{\pm .108}$ & 1.570$^{\pm .039}$ \\   
        AttT2M~\cite{zhong2023attt2m} & 0.413$^{\pm .006}$ & 0.632$^{\pm .006}$ & 0.751$^{\pm .006}$ & 0.870$^{\pm .039}$ & 3.039$^{\pm 
        .021}$ & 10.96$^{\pm .123}$ & \textbf{2.281}$^{\pm .047}$ \\ 
        MoMask~\cite{guo2024momask} & \underline{0.433}$^{\pm .007}$ & \underline{0.656}$^{\pm .005}$ & \underline{0.781}$^{\pm .005}$ & \underline{0.204}$^{\pm .011}$ & \underline{2.779}$^{\pm 
        .022}$ & - & 1.131$^{\pm .043}$ \\  
        \midrule
        T2M Mamba (Ours) & \textbf{0.437}$^{\pm .006}$ & \textbf{0.659}$^{\pm .006}$ & \textbf{0.784}$^{\pm .006}$ & \textbf{0.195}$^{\pm .021}$ & \textbf{2.756}$^{\pm .018}$ & 10.92$^{\pm .096}$ &  1.512$^{\pm .088}$ \\
        \bottomrule
    \end{tabular}}
    \vspace{-0.5em}
    \caption{Comparison of results on the HumanML3D and KIT-ML test set. The right arrow (→) indicates that values closer to real motions correspond to better performance. Each evaluation was run 20 times, and “±” denotes the 95\% confidence interval. Bold text highlights the best scores, while underlined entries mark the second-best.}
    \label{tab:humanml3d}
    \vspace{-3mm}
\end{table*}

\section{ Experiments}
\subsection{Datasets}
\paragraph{Datasets} To evaluate the effectiveness of our method, we conduct experiments on two widely-used text-to-motion benchmarks. (1) HumanML3D ~\cite{guo2022generating}, contains 14,616 human motion sequences paired with 44,970 textual descriptions, offering a diverse and large-scale dataset for text-driven motion generation. (2) KIT-ML ~\cite{plappert2016kit}, introduced by Plappert et al., consists of 3,911 motion sequences annotated with 6,278 natural language descriptions, serving as a standard benchmark for evaluating motion synthesis models.

\begin{figure*}[t] % t = 页面顶部
    \centering
    % 先在图形软件中裁剪，避免在 includegraphics 中使用 trim/clip
    \includegraphics[width=0.78\textwidth]{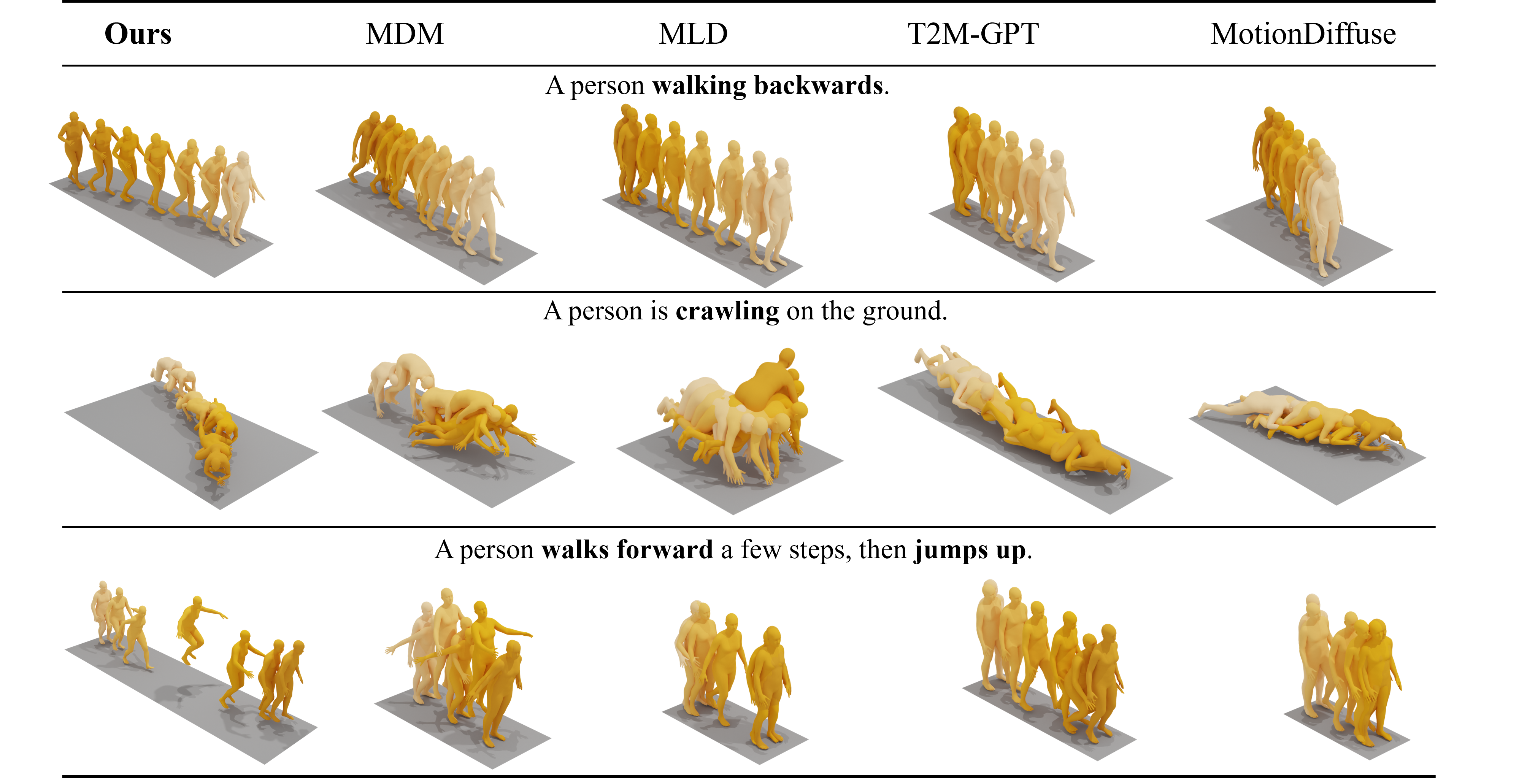}
    \vspace{-0.5em}
    \caption{Qualitative comparison of prominent state-of-the-art methods.}
    \label{motion}
    \vspace{-4mm}
\end{figure*}

\subsection{Comparison with State-of-the-arts}
We evaluate T2M Mamba against state-of-the-art text-to-motion methods on HumanML3D and KIT-ML benchmarks, with results in Table~\ref{tab:humanml3d}. On HumanML3D, it achieves R Precision of 0.506/0.696/0.793 (Top 1/2/3), outperforming AttT2M~\cite{zhong2023attt2m}, T2M-GPT~\cite{zhang2023t2mgptgeneratinghumanmotion}, and Motion Mamba~\cite{zhang2024motion}, while nearing real motions. FID (0.068) and MM Dist (3.034) show superior fidelity and alignment over most baselines; Diversity (9.497) matches real (9.503), and MModality (2.310) ensures varied outputs. On KIT-ML, T2M Mamba outperforms others on most metrics.

We conducted a qualitative comparison of T2M Mamba with prominent state-of-the-art baselines, including MDM~\cite{tevet2022human}, MLD~\cite{chen2023executing}, T2M-GPT~\cite{zhang2023t2mgptgeneratinghumanmotion} and MotionDiffuse~\cite{zhang2024motiondiffuse}.
As shown in Fig~\ref{motion}, for three diverse textual prompts, each model’s generated motion sequence was rendered side by side. Visual inspection shows that T2M Mamba yields motions that are more semantic. T2M Mamba conforms to the prompts, exhibiting superior physical plausibility and outperforming the competing methods.

\begin{figure*}[!t]           % 仍跨两栏，但放宽到顶部
    \centering
    \includegraphics[width=\textwidth]{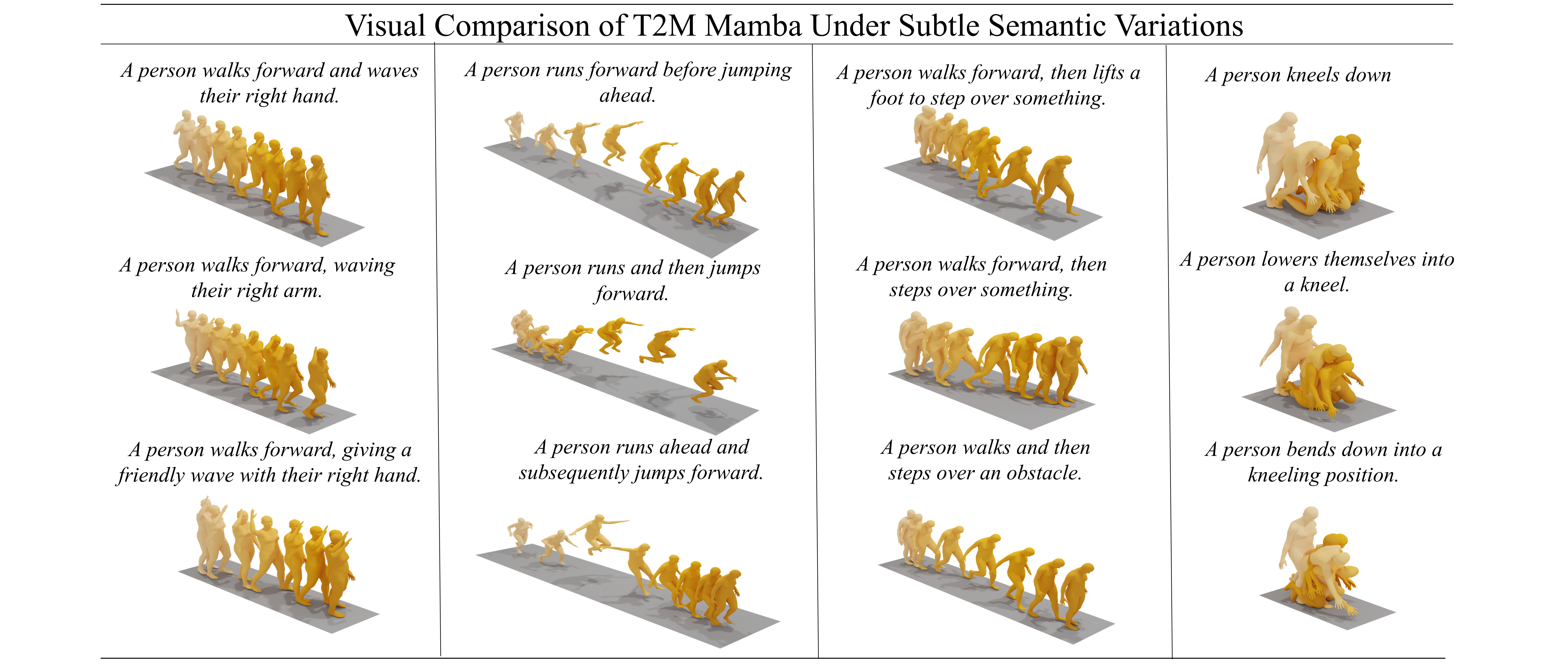}
        \vspace{-5mm}
    \caption{Visual comparison of T2M Mamba under subtle semantic variations}
    \label{fig:motion}
    \vspace{-4mm}
\end{figure*}

\subsection{Ablation Studies}
We conduct ablation studies on the HumanML3D dataset to evaluate the impact of key designs in T2M Mamba, including core modules, the number of T2M Mamba layers, and the effect of both keyframe Weight Estimation and Motion Periodicity Estimation on motion generation quality and computational efficiency. Following the recommendations of MoMask~\cite{guo2024momask}, we adopt FID and R-Precision as the primary evaluation metrics in our ablation study, and additionally report the important metric Multi-Modal Distance. All experiments follow the same evaluation protocol as the main results, with metrics averaged over 20 runs. 
\paragraph{Ablation on Key Modules} To evaluate the effectiveness of our module designs, we perform ablations on the two core components, that is, Periodicity-Saliency Aware Mamba, which integrates keyframe weights $M$ and motion periodicity phase $\mathbf{\Phi}$ to enhance sequence modeling, and Periodic Differential Cross-modal Alignment Module (PDCAM), which reinjects $M$ and $\mathbf{\Phi}$ to achieve robust cross-modal alignment.
% preamble 里确保有：
% \usepackage{graphicx}

\begin{table}[t]
\centering
\small
\label{tab:comparison_ablation}
\resizebox{0.87\columnwidth}{!}{%
\begin{tabular}{l@{\hspace{0.2cm}}c@{\hspace{0.2cm}}c@{\hspace{0.2cm}}c@{\hspace{0.2cm}}c}
\toprule
& R-Top1$\uparrow$ & R-Top3$\uparrow$ & FID$\downarrow$ & MM Dist$\downarrow$\\ 
\midrule
w/o M  & 0.494 & 0.784& 0.088 & 3.122 \\
 
w/o $\phi$      &  0.497& 0.782 & 0.112 & 3.119 \\

w/o M $\&$ $\phi$       & 0.496 & 0.785& 0.108 & 3.112 \\

CroAtten       & 0.456 & 0.755 & 0.131 & 3.240 \\
            
PDCAM(Ours) & \textbf{0.506} & \textbf{0.793} & \textbf{0.068} & \textbf{3.034} \\
\bottomrule
w/o M  & 0.494 & 0.781 & 0.091 & 3.125 \\
            
w/o $\phi$      &  0.499 & 0.787 & 0.079 & 3.092\\
            
w/o M $\&$ $\phi$       & 0.500 & 0.787 & 0.125 & 3.086 \\

w/ M $\&$ $\phi$(Ours) & \textbf{0.506} & \textbf{0.793} & \textbf{0.068} & \textbf{3.034} \\ 
\bottomrule
\end{tabular}%
}
 \caption{Ablation study on T2M Mamba's core modules on HumanML3D. Upper: PDCAM variants; Lower: Periodicity-Saliency Aware Mamba ablations (w/o M: no keyframe weighting; w/o $\phi$: no motion periodicity phase; w/o M $\&$ $\phi$: neither; CroAtten: PDCAM replaced by standard cross-attention). Bold values highlight best performance by full models.}
    \label{core module}
        \vspace{-9mm}
\end{table}

As shown in Table~\ref{core module}, ablating key components in the PDCAM path leads to performance degradation. Removing $M$ increases FID from 0.068 to 0.088 (a 29.4\% rise), with MM Dist rising by 0.088. Removing $\phi$ elevates FID to 0.112 and slightly reduces R-Top1 to 0.497. These results indicate that keyframe learning and periodicity control are interdependent rather than isolated. Joint removal yields an FID of 0.108. Replacing PDCAM with standard cross-attention decreases R-Top3 from 0.793 to 0.755 and increases MM Dist from 3.034 to 3.240, demonstrating that PDCAM substantially enhances cross-modal mapping.

In the Periodicity-Saliency Aware Mamba Module, lacking $M$ raises FID to 0.091, underscoring keyframe weighting's role in mitigating historical forgetting. Lacking $\phi$ increases FID to 0.079, highlighting the importance of periodicity phase for capturing latent motion rhythms. Removing both results in an FID of 0.125 (an 83.8\% degradation). When both $M$ and $\phi$ are retained, the two paths achieve the lowest FID of 0.068 and the highest R-Top3 of 0.793.

\paragraph{Ablation on the Layer Number of T2M Mamba} To assess the effect of model depth, we vary the number of layers in the Mamba backbone from 4 to 10 while keeping other hyperparameters fixed. Results are presented in Table~\ref{layers}. Performance improves with increasing depth up to 6 layers, achieving the best scores across most metrics. Beyond 6 layers, metrics degrade slightly, with FID rising to 0.091 at 7 layers and R-Top3 dropping to 0.783. This suggests that 6 layers strike an optimal balance between capacity and overfitting, as deeper models may introduce redundancy without proportional gains in sequence modeling.
\paragraph{Analysis of Keyframe Weight Estimation and Motion Periodicity Estimation}
To assess the impact of keyframe weight estimation and motion periodicity estimation, we ablate by removing all related components, converting Periodicity-Saliency Aware Mamba to standard bidirectional Mamba without keyframe weighting or periodicity injection, and PDCAM to plain differential attention without keyframe saliency and phase modulation. As shown in Table~\ref{M}, this raises FID from 0.068 to 0.139 (104.4\% increase) and drops R-Top3 from 0.793 to 0.775. These modules incur negligible overhead (AITS increases from 0.40 to 0.41, with unchanged model parameters), yet significantly enhance motion quality and stability via explicit keyframe attention and periodic rhythm learning. periodic rhythms.

\paragraph{Paraphrase Robustness Enabled by PDCAM}
As shown in Figure~\ref{fig:motion}, we prompt an LLM to produce four pairs of action descriptions, each pair containing three sentences with minor semantic variations, and feed them to T2M Mamba to synthesize motions. The results show our model accurately follows the descriptions, and such small changes do not cause divergent or erroneous motions. This indicates that T2M Mamba overcomes the paraphrase sensitivity of prior methods, where even minor synonym substitutions can induce embedding drift and decoding errors.

\begin{table}[t]
\small
\centering
\label{tab:comparison_ablation2}
\resizebox{0.87\columnwidth}{!}{%
\begin{tabular}{l@{\hspace{0.2cm}}c@{\hspace{0.2cm}}c@{\hspace{0.2cm}}c@{\hspace{0.2cm}}c}
% \centering
\toprule
Layer Number N & R-Top1$\uparrow$ & R-Top3$\uparrow$ & FID$\downarrow$ & MM Dist$\downarrow$\\ 
\midrule
\hspace{12mm}4        & 0.498 & 0.791 & 0.089 & 3.074  \\
\hspace{12mm}5        & 0.497 & 0.789& 0.095 & 3.081  \\
\hspace{12mm}6       & \textbf{0.506}&  \textbf{0.793} & \textbf{0.068} & \textbf{3.034}  \\
\hspace{12mm}7       & 0.497 & 0.783 & 0.091 & 3.101 \\
\hspace{12mm}8        & 0.497& 0.779 & 0.088 & 3.126 \\
\hspace{11mm}10        & 0.494 & 0.782 & 0.137 & 3.113\\
\bottomrule
\end{tabular}
}
\vspace{-0.5em}
    \caption{The effect of Layer Number on T2M Mamba Performance on HumanML3D. Bold
values highlight the best performance achieved by our full
models.}
    \label{layers}
           \vspace{-9mm}
  
\end{table}

\begin{table}[t]
\small
\centering
\label{tab:comparison_ablation}
\resizebox{0.80\columnwidth}{!}{%
\begin{tabular}{l@{\hspace{0.2cm}}c@{\hspace{0.2cm}}c@{\hspace{0.2cm}}c@{\hspace{0.2cm}}c}
\toprule
 & \#Params  &AITS $\downarrow$ & R-Top3$\uparrow$ & FID$\downarrow$\\
\midrule
w/o $\phi$ & 24.6M & 0.41 & 0.782 & 0.098 \\
w/o M & 24.6M & 0.41 & 0.785 & 0.134 \\
w/o M $\&$ $\phi$ & 24.6M & \textbf{0.40} & 0.775 & 0.139 \\
w/  M $\&$ $\phi$(Ours) & 24.6M & 0.41 & \textbf{0.793} & \textbf{0.068} \\
\bottomrule
\end{tabular}}
    \caption{Ablation study on keyframe weighting ($M$) and motion periodicity phase ($\phi$) in T2M Mamba. AITS is the inference time and bold values highlight the best performance.}
    \label{M}
     \vspace{-8mm}
\end{table}

\subsection{Implementation Details}
All experiments are conducted on a single NVIDIA RTX 4090 GPU. Training uses a linear beta schedule with 1000 diffusion steps, 140,000 iterations, AdamW optimizer~\cite{loshchilov2017decoupled} (LR $2 \times 10^{-4}$, weight decay $10^{-2}$, gradient clip norm 1), and batch size 128; LR decays by 0.9 every 5,000 steps. Classifier-free guidance masks conditioning with 0.1 probability. Inference employs UniPC~\cite{zhao2023unipc} with 10 timesteps. Average training time is 20.5 hours.

\section{Conclusion}

In this work, we introduced T2M Mamba, a novel framework that couples motion periodicity and keyframe saliency via Periodicity-Saliency Aware Mamba and PDCAM to enable stable text-driven motion generation. By mitigating historical forgetting, capturing latent rhythms, and ensuring robust cross-modal alignment, our approach achieves significant performance on HumanML3D and KIT-ML datasets, with an FID of 0.068 and consistent gains across metrics. Future efforts will explore extensions to real-time applications.

\bibliographystyle{IEEEtran}
\bibliography{IEEEabrv, ICRA}

\end{document}